\documentclass{article}

\usepackage[nonatbib,final]{neurips_2024}

\usepackage[utf8]{inputenc} 
\usepackage[T1]{fontenc}    
\usepackage{hyperref}       
\usepackage{url}            
\usepackage{booktabs}       
\usepackage{amsfonts}       
\usepackage{nicefrac}       
\usepackage{microtype}      
\usepackage{xcolor}         
\usepackage{pifont}
\usepackage{wrapfig}

\usepackage{times}
\usepackage{latexsym}
\usepackage{amssymb}
\usepackage{amsmath}
\usepackage{enumitem}
\usepackage{float}
\usepackage{booktabs}
\usepackage{tablefootnote}
\usepackage{tabularray}

\usepackage{inconsolata}

\usepackage{graphicx}

\title{SciDFM: A Large Language Model with Mixture-of-Experts for Science}

\author{
 Liangtai Sun\textsuperscript{1} \And
 Danyu Luo\textsuperscript{1} \And
 Da Ma\textsuperscript{1} \And
 Zihan Zhao\textsuperscript{1} \And
 Baocai Chen\textsuperscript{1} \AND
 Zhennan Shen\textsuperscript{1} \And
 Su Zhu\textsuperscript{3} \And
 Lu Chen\textsuperscript{1,2}\thanks{The corresponding authors are Lu Chen and Kai Yu.} \And
 Xin Chen\textsuperscript{2} \And
 Kai Yu\textsuperscript{1,2}\footnotemark[1] \and \\
 \textsuperscript{1}X-LANCE Lab, Department of Computer Science and Engineering\\MoE Key Lab of Artificial Intelligence, SJTU AI Institute\\
Shanghai Jiao Tong University, Shanghai, China \\
 \textsuperscript{2}Suzhou Laboratory, Suzhou, China \\
 \textsuperscript{3}AI Speech Co, .Ltd., Suzhou, China \\
 \{slt19990817, chenlusz, kai.yu\}@sjtu.edu.cn
}

\begin{document}

\maketitle

\begin{abstract}

Recently, there has been a significant upsurge of interest in leveraging large language models (LLMs) to assist scientific discovery. However, most LLMs only focus on general science, while they lack domain-specific knowledge, such as chemical molecules and amino acid sequences. To bridge these gaps, we introduce SciDFM, a mixture-of-experts LLM, which is trained from scratch and is able to conduct college-level scientific reasoning and understand molecules and amino acid sequences. We collect a large-scale training corpus containing numerous scientific papers and books from different disciplines as well as data from domain-specific databases. We further fine-tune the pre-trained model on lots of instruction data to improve performances on downstream benchmarks. From experiment results, we show that SciDFM achieves strong performance on general scientific benchmarks such as SciEval and SciQ, and it reaches a SOTA performance on domain-specific benchmarks among models of similar size. We further analyze the expert layers and show that the results of expert selection vary with data from different disciplines. To benefit the broader research community, we open-source SciDFM at \url{https://huggingface.co/OpenDFM/SciDFM-MoE-A5.6B-v1.0}.


\end{abstract}

\section{Introduction}

The advent of Large Language Models (LLMs)~\cite{achiam2023gpt,touvron2023llama,reid2024gemini} has ignited a revolution in the realm of artificial intelligence and has pushed the field of AI for Science (AI4S) to an unprecedented new height. LLMs have demonstrated promising performances in assisting and accelerating scientific discovery~\cite{taylor2022galactica,zhao2024chemdfm,zhao2024chemdfmx,zhang2024sciglm}, such as protein design~\cite{pei2024leveraging}, weather forecasting~\cite{bi2022pangu}, and geoscience~\cite{lin2023geogalactica}.
Despite remarkable achievements in science, LLMs primarily focus on general scientific knowledge represented in text form, ignoring domain-specific contents such as molecules in chemistry and proteins in biology, which are fundamental to advances in these fields.

To overcome this limitation and fully exploit the potential of LLMs for scientific discovery, we introduce SciDFM, a mixture-of-experts Large Language Model trained from scratch with 18.2 billion parameters in total and 5.6 billion parameters activated. SciDFM integrates a mixture-of-experts (MoE)~\cite{lepikhin2020gshard,jiang2024mixtral,rajbhandari2022deepspeed} architecture into a transformer-based~\cite{vaswani2017attention} framework, aiming at enhancing its sophisticated scientific reasoning and understanding capabilities and better modeling similarities and differences across different disciplines and modalities, i.e. text, molecule, protein. In this paper, we detail the pretraining and instruction tuning process of SciDFM, including training corpus and settings. SciDFM leverages a carefully curated corpus of scientific literature and domain-specific databases for pretraining to capture vast scientific knowledge, and is also trained on a large general corpus to retain general knowledge, consuming about 1.1T tokens in total. We meticulously fine-tune SciDFM using a set of instruction-following data containing about 9.3M samples, including interpreting molecular structures and amino acid sequences, thereby improving the performance on downstream benchmarks and bridging the gap in domain-specific knowledge.

To illustrate the prowess of SciDFM, we conduct extensive experiments on several scientific benchmarks. Empirical evaluations affirm the efficacy of SciDFM, demonstrating its superiority on both general scientific benchmarks like SciEval~\cite{sun2024scieval} and SciQ~\cite{welbl2017crowdsourcing}, as well as achieving state-of-the-art (SOTA) performance in domain-specific tasks among models of similar size, such as Mol-Instructions~\cite{fang2023mol}. Our analysis further delves into the results of the expert layer selection, revealing their adaptability to different scientific disciplines, and demonstrating the effectiveness of the MoE model in multidisciplinary scenarios. To benefit the broader research community, we will make SciDFM openly accessible.




\section{SciDFM}

In this section, we introduce the pretraining and instruction tuning details of SciDFM, including the training data construction, model architecture and infrastructure.

\subsection{Pretraining}

\subsubsection{Model Architecture}

\begin{wraptable}{r}{0.36\textwidth}
\centering
\begin{tabular}{lr}
\toprule
Parameter        & Value \\ \midrule
dim              & 3200  \\
n\_layers        & 26    \\
head\_dim        & 100   \\
ffn\_hidden\_dim & 8640  \\
n\_heads         & 32    \\
n\_kv\_heads     & 32    \\
context\_len     & 8192  \\
vocab\_size      & 32192 \\
num\_experts     & 8     \\
topk\_experts    & 2    \\ \bottomrule
\end{tabular}
\caption{Hyper-parameters of SciDFM}
\label{tab:architecture}
\end{wraptable}
SciDFM is based on a transformer architecture~\cite{vaswani2017attention}, and follows modifications of Llama~\cite{touvron2023llama}, i.e. RMSNorm, RoPE and SwiGLU. SciDFM uses the same hyper-parameters of OpenLLaMa-3B~\cite{openlm2023openllama}, the details are shown in Table \ref{tab:architecture}. And in order to better model knowledge of different disciplines, we replace the feed-forward block with Mixture-of-Expert (MoE) layers~\cite{lepikhin2020gshard,jiang2024mixtral}.

We also use the tokenizer of OpenLLaMa-3B, which is trained from scratch using the Bype-Pair Encoding (BPE) method. To better encode the molecules and amino acid sequences and distinguish them from normal characters for better modeling, we treat each chemical atom and amino acid character as a single token and add them into the vocabulary, with special identifiers wrapped~\cite{taylor2022galactica}. For example, molecules C(C(=O)O)N will be encoded as C,(,C,(,=,O,),O,),N, and amino acid sequences MIRLGAPQTL will be encoded as M,I,R,L,G,A,P,Q,T,L, where the special identifiers are omitted.

\subsubsection{Data Construction}

\begin{table*}[!t]
\centering
\begin{tabular}{l|cc}
\toprule
Data Source              & Domain          & \# Tokens (B) \\ \midrule
AMPS~\cite{hendrycks2021measuring}
& Math            & 1.07         \\
OEIS                    & Math            & 0.08         \\
Proof-Pile-2~\cite{azerbayev2023llemma}             & Math            & 25.32         \\
MathGLM-dataset~\cite{yang2023gpt}               & Math            & 9.66          \\
MathPile~\cite{wang2023mathpile}                 & Math            & 0.21         \\
PubChem Compound         & Chemistry       & 18.34         \\
USPTO                    & Chemistry       & 0.13         \\
Uniprot                  & Biology         & 0.52         \\
BioRxiv                  & Biology         & 2.77         \\
MedRxiv                  & Biology         & 0.31         \\
RefSeq Genome            & Biology         & 1.38         \\
Uniref                   & Biology         & 16.85        \\
GeoNames                 & Geography       & 0.82         \\
the-stack~\cite{Kocetkov2022TheStack}                & Science Code    & 9.65     \\
Scientific Papers        & General Science & 182.65        \\
Scientific Books         & General Science & 7.00          \\
SlimPajama-Arxiv~\cite{soboleva2023slimpajama}         & General Science & 29.78         \\
LibreText                & General Science & 0.05          \\ \midrule
\textit{Total Science}            & -               & 306.59        \\ \midrule
WikiBooks                & General         & 0.14          \\
En-Zh-Trans              & General         & 3.10          \\
Wikipedia-Zh             & General         & 0.56          \\
Baike                    & General         & 23.16         \\
SlimPajama(w/o Arxiv)~\cite{soboleva2023slimpajama}    & General         & 241.64|241.75 \\ \midrule
\textit{Total General}            & -               & 268.60|268.71 \\ \bottomrule  
\end{tabular}
\caption{The training corpus details of SciDFM.}
\label{tab:data_details}
\end{table*}

To enhance the understanding and reasoning abilities of SciDFM on science domain, we collect a large-scale training corpus containing a large number of open-access scientific papers and books of different disciplines. And to acquire domain-specific knowledge, we also include data from some databases. Furthermore, in order to maintain the generic capabilities of SciDFM, we use data from some open-source generic corpora. The details of our pretraining corpus are shown in Table \ref{tab:data_details}. Our pretraining corpus contains about 300B science-domain tokens and 270B general-domain tokens, with 570B tokens in total. We train SciDFM for two epochs, and for the second epoch, we re-sample data of C4, CC, and Github subsets of SlimPajama~\cite{soboleva2023slimpajama,together2023redpajama}. And for the-stack dataset, we only use data from programming languages that are relevant to scientific computing, such as Matlab and Scilab.


\begin{table*}[!ht]
\centering
\begin{tabular}{l|cc}
\toprule
Data             & Domain                   & \# Samples \\ \midrule
Arxivphy         & Physics                  & 30231      \\
MATH~\cite{hendrycks2021measuring}             & Math                     & 7500       \\
MetaMathQA~\cite{yu2023metamath}       & Math                     & 395000     \\
MathInstruct~\cite{yue2023mammoth}     & Math                     & 262040     \\
OrcaMath~\cite{mitra2024orcamath}         & Math                     & 200035     \\
DeepLoc2~\cite{thumuluri2022deeploc}         & Biology                  & 22642      \\
BioASQ~\cite{tsatsaronis2015overview}           & Biology                  & 8021       \\
MedMCQA~\cite{pmlr-v174-pal22a}          & Biology                  & 182822     \\
MedQA~\cite{jin2021disease}            & Biology                  & 10178      \\
PubMedQA~\cite{jin2019pubmedqa}         & Biology                  & 500        \\
Mol-Instructions~\cite{fang2023mol} & Chemistry \& Biology     & 1863630    \\
ChemDFM-sft~\cite{zhao2024chemdfm}      & Chemistry                & 1818154    \\
SciQ~\cite{welbl2017crowdsourcing}             & General Science          & 11679      \\
Camel~\cite{li2023camel}            & General Science          & 110000     \\
SciInstruct~\cite{zhang2024sciglm}      & General Science          & 82640      \\
WebInstructSub~\cite{yue2024mammoth2}   & General Science                  & 2327291    \\
MMLU~\cite{hendryckstest2021}             & General                  & 99842      \\
LIMA~\cite{zhou2024lima}             & General                  & 1000       \\
ROPES~\cite{Lin2019ReasoningOP}            & General                  & 10924      \\
QASC~\cite{allenai:qasc}             & General                  & 8134       \\
OpenBookQA~\cite{OpenBookQA2018}       & General                  & 4957       \\
Dolly~\cite{conover2023free}            & General                  & 15011      \\
SlimOrca-Dedup~\cite{SlimOrcaDedup}   & General                  & 363491     \\
GPT4All~\cite{anand2023gpt4all}          & General                  & 806199     \\
Other            & General                  & 994896     \\ \midrule
\textit{Total(after dedup)} & -                      & 9325310    \\ \bottomrule
\end{tabular}
\caption{Details of instruction tuning dataset.}
\label{tab:it_data}
\end{table*}

\begin{table*}[!ht]
\centering
\tabcolsep=5pt
\begin{tabular}{l|ccc|cc|ccc|c}
\toprule
Model              & \ding{172} & \ding{173}  & \ding{174} & \ding{175} & \ding{176}  & \ding{177} & \ding{178} & \ding{179} & Avg.   \\ \midrule
Llama2-7B          & 27.06   & 57.00 & 36.43/46.59  & 3.94  & 3.96  & 26.32 & 29.84   & 66.80    & 32.95 \\
Galactica-6.7B     & 46.28   & 74.20 & 44.28/61.83  & 2.80  & 6.32  & 30.48 & 36.46   & 48.80    & 38.91 \\
Llama2-13B         & 33.88   & 78.10 & 56.66/72.35  & 22.82 & 3.90  & 32.68 & 34.28   & 77.80    & 45.45 \\
ChatGLM2-6B        & 54.25   & 75.80 & 57.08/73.57  & 25.09 & 7.18  & 27.42 & 34.21   & 60.40    & 45.94 \\
Galactica-30B      & 54.24   & 83.10 & 57.85/75.04  & 13.65 & 8.66  & 37.71 & 48.43   & 58.80    & 48.35 \\
Llama3-8B          & 59.70   & 90.00 & 71.16/84.05  & 5.91  & 7.00  & 48.78 & 52.74   & 26.60    & 49.59 \\
ChatGLM3-6B        & 51.13   & 77.60 & 60.84/75.97  & 60.27 & 23.52 & 24.59 & 31.39   & 51.80    & 50.53 \\
SciGLM-6B          & 61.22   & 88.70 & 77.47/86.57  & 42.23 & 16.40 & 42.81 & 44.94   & 73.60    & 59.12 \\
ChatGLM3-6Bb   & 60.34   & 89.00 & \textbf{78.58}/\textbf{87.37}  & 59.82 & 22.64 & 42.73 & 45.14   & 74.40    & 61.96 \\
Llama3-8B-it & \textbf{64.91}   & \textbf{91.60} & 76.45/87.33  & \textbf{76.57} & 26.26 & \textbf{56.48} & \textbf{59.31}   & 72.00    & 67.44 \\ 
SciDFM(ours)             & 62.48   & 88.00 & 64.76/81.48  & 59.14 & \textbf{27.28} & 44.54 & 53.10   & \textbf{78.00}    & 61.56 \\ \bottomrule
\end{tabular}
\caption{Main Results on 8 general scientific language understanding and reasoning tasks. ChatGLM3-6Bb stands for \texttt{ChatGLM3-6B-base}, Llama3-8B-it stands for \texttt{Llama3-8B-Instruct}. The columns from \ding{172} to \ding{179} stand for SciEval, SciQ, ARC, GSM8K, MATH, MedQA, MedMCQA. PubMedQA, respectively, where the results of ARC are shown in the form of ARC\_C/ARC\_E.}.
\label{tab:lu_eval_res}
\end{table*}

\subsubsection{Training Details}

Following Llama~\cite{touvron2023llama}, SciDFM is trained from scratch using the AdamW optimizer~\cite{loshchilov2017decoupled} with $\beta_1=0.9, \beta_2=0.95$. We use a cosine learning rate schedule, such that the final learning rate is equal to 10\% of the initial learning rate. We use a weight decay of 0.1 and gradient clipping of 1.0, and we set the macro batch size to 4M tokens and sequence length to 8192 tokens. For MoE layers, we set the auxiliary loss factor to 0.02 and set the expert capacity factor to 1.0. In total, we train SciDFM for two epochs, resulting in about 1.1T tokens fed. We use a learning rate of $3e-4$ for the first epoch and $3e-5$ for the second epoch, while the other settings remain the same as the above. We train SciDFM on a cluster of 16 nodes, with 8 A800 GPUs on each node for about two months.

\subsection{Instruction Tuning}

To improve the performance of SciDFM on downstream benchmarks, we collect a number of instruction-tuning data from open-source datasets. The details are shown in Table \ref{tab:it_data}. We fine-tune the pre-trained SciDFM for 5 epochs in total. During fine-tuning, we use a learning rate of 2e-5, and we set the sequence length to 2048 and macro batch size to 32. The other settings are the same as the pretraining stage.

\section{Evaluation}

In this section, we show the performance of SciDFM on some general and domain-specific benchmarks, and analyze the results of expert selection in different domains.

\subsection{Evaluation Setup}

\begin{table*}[!ht]
\centering
\begin{tabular}{l|cc}
\toprule
Dataset          & \# Samples & Subject \\ \midrule
SciEval          & 15,901      &  $\diamondsuit,\triangle,\square,\heartsuit$       \\
SciQ             & 1,000       &  $\diamondsuit,\triangle,\square,\heartsuit$       \\
ARC\_C           & 1,172       &  $\diamondsuit,\triangle,\square,\heartsuit$       \\
ARC\_E           & 2,376       &  $\diamondsuit,\triangle,\square,\heartsuit$       \\
GSM8K            & 1,319         &  $\square$       \\
MATH             & 5,000         &  $\square$       \\
MedQA            & 1,273       &  $\triangle$       \\
MedMCQA          & 4,183       &  $\triangle$       \\
PubMedQA         & 500         &  $\triangle$       \\
Mol-Instructions & 24,324        &  $\diamondsuit,\triangle$       \\
MoleculeNet      & 11,767      &  $\diamondsuit$       \\ \bottomrule
\end{tabular}
\caption{Overview of evaluation benchmarks. ($\diamondsuit$: Chemistry, $\triangle$: Biology, $\square$: Math, $\heartsuit$: Physics)}
\label{tab:eval_dataset}
\end{table*}

\begin{table*}[]
\centering
\begin{tabular}{l|ccc}
\toprule
Model              & Avg.G & Avg.M & Avg.B \\ \midrule
LLaMa2-7B          & 41.77        & 3.95      & 40.99    \\
Galactica-6.7B     & 56.65        & 4.56      & 38.58    \\
LLaMa2-13B         & 60.25        & 13.36     & 48.25    \\
ChatGLM2-6B        & 65.18        & 16.14     & 40.68    \\
Galactica-30B      & 67.56        & 11.16     & 48.31    \\
LLaMa3-8B          & 76.23        & 6.46      & 42.71    \\
ChatGLM3-6B        & 66.39        & 41.90     & 35.93    \\
SciGLM-6B          & 78.49        & 29.32     & 53.78    \\
ChatGLM3-6B-base   & 78.82        & 41.23     & 54.09    \\
Llama3-8B-Instruct & 80.07        & 51.42     & 62.60    \\
SciDFM(ours)             & 74.18        & 43.21     & 58.55    \\ \bottomrule
\end{tabular}
\caption{Average Results of General Science, Math and Biology tasks on scientific language understanding and reasoning tasks. Avg.G, Avg.M and Avg.B stand for average performance on general science, math and biology tasks respectively.}
\label{tab:avg_eval_res}
\end{table*}

\paragraph{Evaluation Tasks} Summarization of evaluation tasks are shown in Table \ref{tab:eval_dataset}. These evaluation datasets cover a wide range of subjects, including math, chemistry, physics, biology, protein and molecule.
\begin{itemize}[itemsep=0pt,leftmargin=1em]
    \item SciEval~\cite{sun2024scieval} is a comprehensive and multidisciplinary benchmark designed to assess the scientific research capabilities of Large Language Models.
    \item SciQ~\cite{welbl2017crowdsourcing} comprises high-quality, domain-targeted multiple-choice science exam questions, created through a novel method that combines crowd-sourcing with AI-assisted document and answer option selection.
    \item ARC~\cite{clark2018think} presents a sophisticated question set, designed to advance AI research in complex question answering within the context of grade-school science, composed of an easy subset and a challenging subset.
    \item GSM8K~\cite{cobbe2021training} is composed of diverse and linguistically rich grade school math word problems, designed to benchmark and improve the multi-step mathematical reasoning abilities of LLMs, revealing their limitations in handling complex reasoning tasks.
    \item MATH~\cite{hendrycks2021measuring} contains challenging competition mathematics problems, each with detailed solutions, designed to assess and enhance mathematical problem-solving and reasoning capabilities of LLMs, accompanied by an auxiliary pretraining dataset to bolster fundamental math understanding.
    \item MedQA~\cite{jin2021disease} is a pioneering multiple-choice dataset for open-domain question answering in the medical field, encompassing a number of questions sourced from professional medical board exams.
    \item MedMCQA~\cite{pmlr-v174-pal22a} is a large-scale medical multiple-choice question-answering dataset, spanning 2,400 healthcare topics and 21 subjects, designed to challenge models with diverse reasoning skills across various medical domains.
    \item PubMedQA~\cite{jin2019pubmedqa} is a biomedical question-answering dataset based on PubMed abstracts, requiring quantitative reasoning over research texts.
    \item Mol-Instructions~\cite{fang2023mol} is a specialized, meticulously curated dataset containing diverse biomolecular instructions, designed to enhance LLMs' understanding and prediction capabilities within the realms of molecular, protein, and broader biomolecular texts.
    \item MoleculeNet~\cite{wu2018moleculenet} is a comprehensive benchmark dataset for molecular machine learning, featuring curated public datasets, standardized evaluation metrics, and open-source implementation of various molecular featurization and learning methods.
\end{itemize}

In our experiments, we take Mol-Instructions and MoleculeNet as domain-specific tasks, and take the remaining benchmarks as general scientific language understanding and reasoning tasks.


\begin{table*}[!ht]
\centering
\begin{tabular}{l|cccccc}
\toprule
Model          & bace & bbbp & ClinTox & HIV & Tox21 & Avg \\ \midrule
LLaMa2-13B-chat     & 26.0 & 60.3 & 45.7 & 29.0 & 51.7 & 42.54 \\
GPT4o(0513)           & 62.5 & 61.5 & 51.6 & 65.9 & 55.2 & 59.34 \\
Galactica-30B  & 72.7 & 59.6 & 82.2 & 75.9 & 68.5 & 71.78 \\
ChemDFM-13B    & \textbf{78.4} & \textbf{66.7} & 89.9 & \textbf{73.6} & \textbf{79.8} & 77.68 \\
SciDFM(ours)         & 76.4 & 64.8 & \textbf{98.5} & 71.5 & 75.6 & 77.36 \\ \bottomrule
\end{tabular}
\caption{The Results of molecular property prediction tasks on MoleculeNet in AUC-ROC scores.}
\label{tab:molecule_res}
\end{table*}

\begin{table*}[t]
\centering
\tabcolsep=4pt
\begin{tabular}{l|cccccc}
\toprule
Model          & Exact$\uparrow$ & Levenshtein$\downarrow$ & RDK FTS$\uparrow$ & MACCS FTS$\uparrow$ & Morgan FTS$\uparrow$ & Validity$\uparrow$ \\ \midrule
\multicolumn{7}{l}{\textit{Description Guided Molecule Design}} \\ \midrule
ChatGLM3-6B      & 0.001 & 199.47 & 0.103 & 0.174 & 0.059 & 0.236 \\
Galactica-6.7B        & 0.000 & 44.152 & 0.134 & 0.248 & 0.088 & 0.992 \\
Text+Chem T5     & \textbf{0.097} & 41.819 & 0.352 & 0.474 & 0.353 & 0.721 \\
Mol-Instructions & 0.002 & \textbf{41.367} & 0.231 & 0.412 & 0.147 & \textbf{1.000} \\
SciDFM(ours)     & 0.084 & 43.414 & \textbf{0.586} & \textbf{0.704} & \textbf{0.443} & 0.994 \\ \midrule
\multicolumn{7}{l}{\textit{Reagent Prediction}} \\ \midrule
ChatGLM3-6B      & 0.000 & 154.73 & 0.044 & 0.144 & 0.046 & 0.275 \\
Galactica-6.7B        & 0.000 & 35.021 & 0.156 & 0.257 & 0.097 & 0.946 \\
Text+Chem T5     & \textbf{0.239} & 20.413 & \textbf{0.705} & \textbf{0.789} & \textbf{0.652} & 0.762 \\
Mol-Instructions & 0.045 & 27.262 & 0.313 & 0.509 & 0.262 & \textbf{1.000} \\
SciDFM(ours)           & 0.192 & \textbf{17.527} & 0.476 & 0.576 & 0.452 & 0.999 \\ \midrule
\multicolumn{7}{l}{\textit{Retrosynthesis}} \\ \midrule
ChatGLM3-6B      & 0.000 & 59.062 & 0.636 & 0.695 & 0.570 & 0.492 \\
Galactica-6.7B         & 0.000 & 30.760 & 0.036 & 0.127 & 0.051 & 0.995 \\
Text+Chem T5     & 0.000 & 49.323 & 0.039 & 0.186 & 0.052 & 0.313 \\
Mol-Instructions & 0.044 & 23.167 & 0.237 & 0.364 & 0.213 & \textbf{1.000} \\
SciDFM(ours)           & \textbf{0.665} &  \textbf{6.45}  & \textbf{0.916} & \textbf{0.937} & \textbf{0.888} & 0.998
\\\bottomrule
\end{tabular}
\caption{The Results of molecular generation tasks on Mol-Instructions. Exact stands for exact matches, and validity stands for valid molecules. RDK, MACCS and Morgan are three kinds of molecular fingerprints.}
\label{tab:molins_mol_res}
\end{table*}

\paragraph{Evaluation Methods} 

Since SciDFM is an instruction-following model by default, we conduct all experiments using zero-shot settings. And most of the models we select for comparison are able to follow instructions: 
\begin{itemize}[itemsep=0pt,leftmargin=1em]
    \item Galactica~\cite{taylor2022galactica} is a large language model specifically designed to store, combine, and reason about vast amounts of scientific knowledge, outperforming existing models on various scientific tasks and aiming to serve as a new, advanced interface for scientific research. We select \texttt{Galactica-30B} and \texttt{Galactica-6.7B} for comparison.
    \item Llama~\cite{touvron2023llama} is a series of open-source powerful language models, ranging from 7 billion to 70 billion parameters, trained on massive public datasets, and outperforms many of the available open-source models on common benchmarks. We select \texttt{Llama3-8B}, \texttt{Llama3-8B-Instruct}, \texttt{Llama2-7B} and \texttt{Llama2-13B} for comparison.
    \item ChatGLM~\cite{zeng2022glm,du2022glm} is a series of advanced language models, excel in various metrics and tasks, rivaling or surpassing counterparts like GPT-4, thanks to their extensive training on multilingual data, specialized alignment techniques, and the ability to integrate diverse tools dynamically. We select \texttt{ChatGLM2-6B}, \texttt{ChatGLM3-6B} and \texttt{ChatGLM3-6B-base} for comparison.
    \item SciGLM~\cite{zhang2024sciglm} is a suite of scientific language models that enhance college-level scientific reasoning through a self-reflective instruction annotation framework, addressing data scarcity in the science domain, and improving upon ChatGLM in handling complex scientific and mathematical problems without compromising language understanding. We select \texttt{SciGLM-6B} for comparison.
    \item ChemDFM~\cite{zhao2024chemdfm} is specifically trained for Chemistry, combining knowledge from chemical literature and general domains to excel in understanding, reasoning, and applying chemical information, outperforming generic LLMs and even GPT-4 on chemical tasks. We select \texttt{ChemDFM-13B} for comparison.
\end{itemize}

\subsection{Main Results}

\paragraph{General Scientific Benchmark}
Table \ref{tab:lu_eval_res} presents the evaluation results on eight general scientific language understanding and reasoning tasks. The results show that SciDFM reaches a better performance on average than Galactica-series models, Llama-series models except \texttt{Llama3-8B-Instruct} and ChatGLM-series models except \texttt{ChatGLM3-6B-base}. In Table \ref{tab:avg_eval_res}, we also present the average performance of general science, math and biology tasks on the above eight benchmarks, in which SciEval, SciQ and ARC belong to general science task, GSM8K and MATH belong to math task, MedQA, MedMCQA and PubMedQA belong to biology task. We find that SciDFM outperforms all models except \texttt{Llama3-8B-Instruct} on math and biology domain, while it is weak in general science tasks. In conclusion, SciDFM can reach a similar performance to top-tier models of similar amount of compute, while it is weaker compared to models that are larger and trained using more data.

\paragraph{Domain-specific Scientific Benchmark}
Table \ref{tab:molecule_res} presents the performance of molecular property prediction tasks on MoleculeNet. From the results shown in AUC-ROC scores, we find that SciDFM outperforms most LLMs except \texttt{ChemDFM-13B}. To further evaluate model's ability on domain-specific benchmarks, we present the performance of molecule and protein understanding tasks on Mol-Instructions in Table \ref{tab:molins_mol_res} and Table \ref{tab:molins_prot_res}. We find that SciDFM can reach a SOTA performance on molecule and protein understanding tasks.

\begin{table*}[]
\centering
\tabcolsep=4pt
\begin{tabular}{l|cccc}
\toprule
Model          & Protein Function & Functional Description & Catalytic Activity & Domain/Motif \\ \midrule
GPT4o(0513)            & 0.06 & 0.05 & 0.07 & 0.06 \\
ChatGLM         & 0.15 & 0.14 & 0.13 & 0.10 \\
Mol-Instruction & 0.43 & 0.44 & 0.52 & 0.46 \\
SciDFM(ours)          & \textbf{0.60} & \textbf{0.72} & \textbf{0.76} & \textbf{0.55} 
\\\bottomrule
\end{tabular}
\caption{The Results of protein understanding tasks on Mol-Instructions. All tasks are evaluated in ROUGE-L score.}
\label{tab:molins_prot_res}
\end{table*}


\subsection{Expert Choices Analysis}

In this subsection, we conduct analysis on expert choice results on data from different domains. Formally, we denote the output of the $i$th attention layer as $h_i \in \mathbb{R}^{l \times d}$, where $l$ is the sequence length and $d$ is the hidden dimension, and we denote the weight of the $i$th gate network in the corresponding MoE layer as $W_g \in \mathbb{R}^{d\times e}$, where $e$ represents the number of experts. Then, we have $g_i = h_i \cdot W_g \in \mathbb{R}^{l \times e}$, representing the probability of each token being assigned to each expert. Suppose the number of hidden layers is $N$, for a given text $T$, we define the expert choice results as:
\begin{gather}
    e_i = \text{Softmax}(\sum_{j=1}^lg_i[j,:]) \in \mathbb{R}^{e}, \\
    E_T = \text{Concat}([e_1, e_2, \dots, e_N]) \in \mathbb{R}^{Ne}.
\end{gather}

We randomly select 100 research papers each in the fields of math, chemistry, biology, and physics, and also select 100 chemical molecules and 100 amino acid sequences as analysis data. For each text $T$, we calculate $E_T$ using the above formulas. Then, we use the t-SNE~\cite{van2008visualizing} algorithm to reduce them to three dimensions and visualize them. 

\begin{figure}
    \centering
    \includegraphics[width=0.48\textwidth]{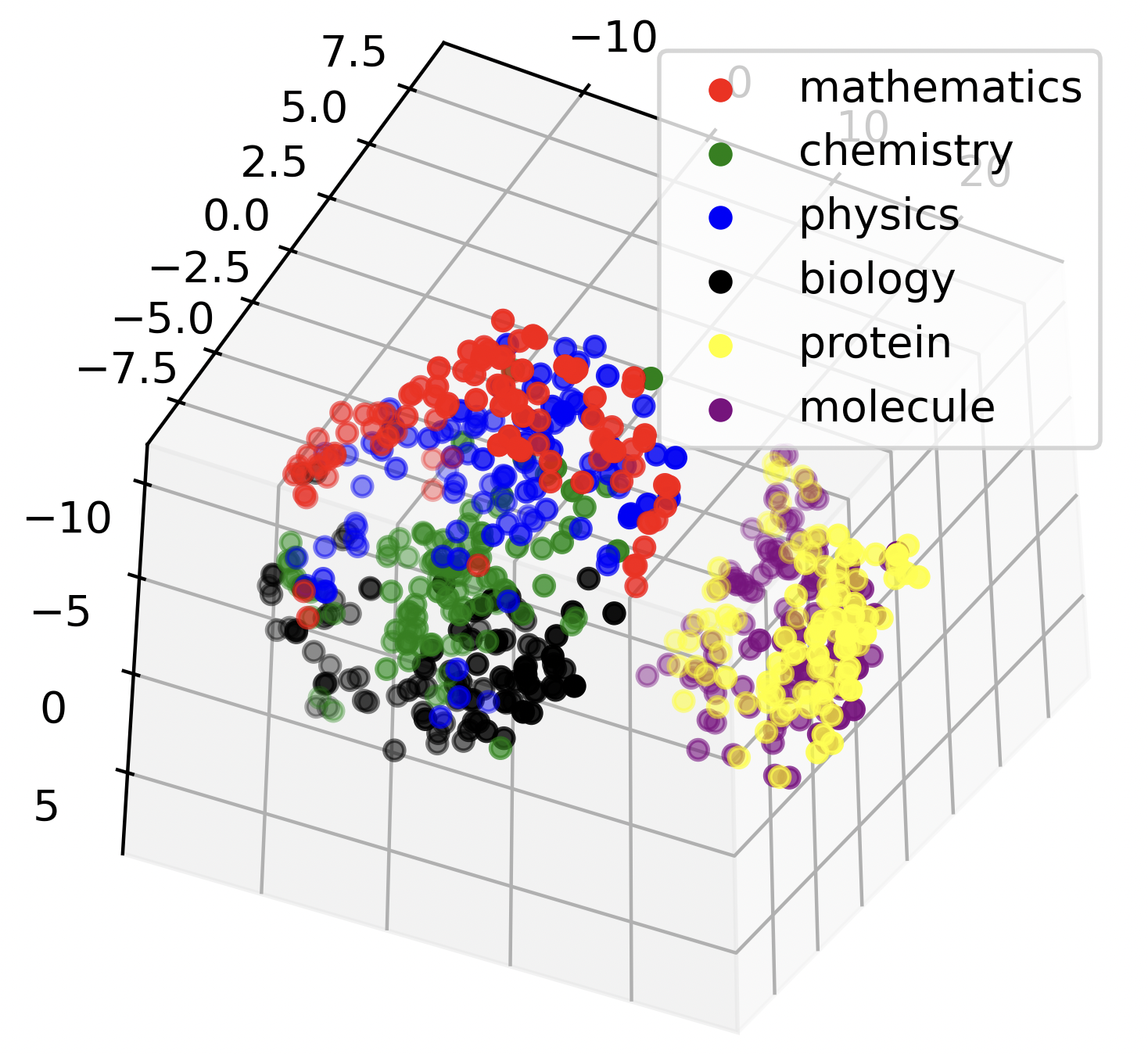}
    \caption{Visualization of t-SNE results using data from different domains.}
    \label{fig:visual}
\end{figure}

The visualization result is shown in Figure \ref{fig:visual}.
It can be found that research papers in mathematics, chemistry, physics, and biology demonstrate a clear clustering pattern, indicative of discipline-specific language characteristics, while chemical molecules and amino acid sequences do not exhibit such a clustering phenomenon.
In addition, the separation of molecular and protein data from other categories is stark, likely due to the unique vocabularies inherent to these domains, which are not shared with the remainder of the dataset.


Furthermore, the spatial proximity observed in the visualization adds another layer of insight: the clusters of mathematics and physics are in close proximity, as are those of chemistry and biology, with chemistry also showing affinity towards physics. This aligns well with the interrelationships and overlapping nature of knowledge between these scientific disciplines, as reflected in their linguistic characteristics.

\section{Related Works}

The success of pretraining language models like BERT~\cite{devlin2018bert} and GPT~\cite{radford2018improving} makes researchers wonder whether the language model can bring about improved performance in the field of Science.
\paragraph{Domain-Specific Language Model for Science}
BioGPT~\cite{luo2022biogpt} is a domain-specific generative language model pre-trained on large-scale biomedical literature, which outperforms previous models on six biomedical-related tasks. Based on case studies, the researchers further demonstrated the advantages of BioGPT in generating fluent descriptions for biomedical terms in biomedical literature. 
ProGen2~\cite{nijkamp2023progen2} is a protein language model pre-trained on a corpus of more than one billion protein sequences including genome, metagenome, and immune library databases. ProGen2 shows optimal performance in capturing observed evolutionary sequence distributions, generating new protein sequences, and predicting protein fitness without additional fine-tuning. 
Med-PaLM~\cite{singhal2023large} is a large language model (LLM) designed to provide high-quality answers to medical questions, which is an instruction prompt-tuned version of Flan-PaLM~\cite{chowdhery2023palm} specialized for the medical domain. They reveal limitations of Flan-PaLM in scientific grounding, harm, and bias through evaluation, while Med-PaLM significantly reduces the gap (or even compares favorably) to clinicians on several of these axes, according to both clinicians and lay users. 
MTL-BERT~\cite{zhang2022pushing} proposes to use large-scale pre-training, multi-task learning, and SMILES enumeration to alleviate the data sparsity problem. It mines the rich contextual information in SMILES strings through self-supervised pre-training, and then fine-tunes the pre-trained model simultaneously using multiple downstream tasks. At the same time, it combines SMILES enumeration as a data augmentation strategy to increase data diversity. Experimental results show that MTL-BERT can achieve optimal performance on molecular datasets. 
ChemDFM~\cite{zhao2024chemdfm} is the first LLM towards Chemical General Intelligence (CGI), which is trained on 34B tokens from chemical literature, textbooks, and instructions as well as various data from the general domain. It can store, understand, and reason over chemical knowledge and languages while still possessing advanced free-form language comprehension capabilities. Extensive quantitative evaluation shows that ChemDFM can significantly outperform the representative open-sourced LLMs.

\paragraph{General-domain Language Model for Science}
SciBERT~\cite{beltagy2019scibert} is a pre-trained language model based on the BERT model architecture, which aims to address the lack of high-quality, large-scale labeled scientific data. SciBERT uses a large multi-domain scientific publication corpus for pre-training to improve the performance of downstream scientific benchmarks and has achieved state-of-the-art performance on multiple tasks.
Galactica~\cite{taylor2022galactica} is a large language model that can store, combine and reason about scientific knowledge, which is trained on a large scientific corpus of papers, reference material, knowledge bases and many other sources. Galactica outperforms previous models on a range of scientific tasks and sets a new state-of-the-art on downstream tasks such as PubMedQA and MedMCQA. SciGLM~\cite{zhang2024sciglm} is a suite of scientific language models designed to enhance college-level scientific reasoning. It utilizes a self-reflective instruction annotation framework to address data scarcity in the science domain. SciGLM significantly improves upon ChatGLM by effectively handling complex scientific and mathematical problems, all while maintaining strong language understanding capabilities.

Compared to prior works, SciDFM either can achieve a better performance, or is more generalized. With the utilization of Mixture-of-Experts architecture, SciDFM can better model similarities and differences across different disciplines and modalities and have stronger sophisticated scientific reasoning and understanding capabilities.

\section{Conclusion}

In this paper, we introduce SciDFM, a mixture-of-experts LLM able to conduct college-level scientific reasoning and understand molecules and amino acid sequences. We show the pretraining and instruction-tuning process of SciDFM in detail, including data, architecture and hyper-parameters. We conduct evaluation on eight general scientific language understanding and reasoning tasks and two domain-specific tasks. From the results, we show that SciDFM achieves strong performance on general scientific benchmarks such as SciEval and SciQ, and it reaches a SOTA performance on domain-specific benchmarks among models of similar size. We further analyze the expert choices of MoE layers and show that the results of expert selection vary with data from different disciplines and exhibit clustering phenomena related to their relationships.

\section*{Acknowledgments and Disclosure of Funding}

This work was supported by the National Science and Technology Major Project 2023ZD0120703, the China NSFC Projects (U23B2057, 62106142 and 62120106006), and the Shanghai Municipal Science and Technology Major Project (2021SHZDZX0102).

\bibliographystyle{unsrt}






\end{document}